\documentclass{article}

\usepackage{microtype}
\usepackage{graphicx}  
\usepackage{subcaption}
\usepackage{booktabs} 
\usepackage[utf8]{inputenc}
\DeclareUnicodeCharacter{FB01}{fi}
\usepackage{latexsym,amsfonts,amssymb,amsthm,amsmath}
\usepackage{xcolor}
\usepackage{amsmath} 
\usepackage{cancel}
\usepackage{wasysym}
\usepackage{booktabs}
\usepackage{tabularx}

\usepackage{hyperref}

\usepackage{xspace}

\usepackage[nonatbib, preprint]{neurips_2020}

\usepackage[utf8]{inputenc} 
\usepackage[T1]{fontenc}    
\usepackage{url}            
\usepackage{booktabs}       
\usepackage{amsfonts}       
\usepackage{nicefrac}       
\usepackage{microtype}      

\theoremstyle{plain}
\newtheorem{thm}{Theorem}[section] 

\theoremstyle{definition}
\newtheorem{defn}[thm]{Definition}

\title{Privacy in Deep Learning: A Survey}

\author{
    Fatemehsadat Mireshghallah\textsuperscript{\rm 1}, Mohammadkazem Taram\textsuperscript{\rm 1}, Praneeth Vepakomma\textsuperscript{\rm 2},\\
    \textbf{Abhishek Singh}\textsuperscript{\rm 2}, \textbf{Ramesh Raskar}\textsuperscript{\rm 2}, \textbf{Hadi Esmaeilzadeh}\textsuperscript{\rm 1},\\
    \textsuperscript{\rm 1} University of California San Diego,
    \textsuperscript{\rm 2} Massachusetts Institute of Technology\\
    \texttt{ \{fatemeh, mtaram, hadi\}@ucsd.edu},\\ \texttt{ \{vepakom, abhi24, raskar\}@mit.edu}
}

\begin{document}

\maketitle
\begin{abstract}
The ever-growing advances of deep learning in many areas including vision, recommendation systems, natural language processing, etc., have led to the adoption of Deep Neural Networks (DNNs) in production systems.
The availability of large datasets and high computational power are the main contributors to these advances. The datasets are usually crowdsourced and may contain sensitive information. This poses serious privacy concerns as this data can be misused or leaked through various vulnerabilities. 
Even if the cloud provider and the communication link is trusted, there are still threats of inference attacks where an attacker could speculate properties of the data used for training, or find the underlying model architecture and parameters. 
In this survey, we review the privacy concerns brought by deep learning, and the mitigating techniques introduced to tackle these issues. 
We also show that there is a gap in the literature regarding test-time inference privacy, and propose possible future research directions.

\end{abstract}
\vspace{-3ex}
\section{Introduction}
\label{sec:intro}
%
%
The success of Deep Neural Networks (DNNs) in various fields including vision, medicine, recommendation systems, natural language processing, etc., has resulted in their deployment in numerous production systems~\cite{nvidia-product, nlp-product, kaissis2020secure, Singh2020BenchmarkingDP}. 
In the world of medicine, learning is used to find patterns in patient histories and to recognize abnormalities in medical imaging which help with disease diagnosis and prognosis. The use of machine learning in healthcare can compromise patient privacy, for instance by exposing the patient's genetic markers, as shown by Fredrikson et al.~\cite{pharmacogenetics}.
Deep learning is also widely used in finance for predicting prices or creating portfolios, among many other applications. In these cases, usually, an entity trains its own model and the model parameters are considered confidential. Being able to find or infer them is considered a breach of privacy~\cite{financial}.
Ease of access to large datasets and high computational power (GPUs and TPUs) have paved the way for the aforementioned advances. These datasets are usually crowdsourced and might contain sensitive information. This poses serious privacy concerns, as neural networks are used in different aspects of our lives~\cite{nytimes2, pharmacogenetics, NYTImes, facebook,google-street, Schiff2009RespectfulCD, 2005CV}.

Figure~\ref{fig:threats} shows a classification of possible threats to deep learning. One threat is the direct intentional or unintentional exposure of sensitive information, through untrusted data curator, communication link, or cloud~\cite{breach1, breach2}.  This information can be the training data, inference queries or model parameters or hyperparameters. If we assume that information cannot be attained directly, there is still the threat of information exposure through inference, indirectly. 
Membership inference attacks~\cite{memberinf} can infer whether a given data instance was part of the training process of a model. 
Model inversion and attribute inference attacks can infer sensitive features about a data instance, from observed predictions of a trained model, and other non-sensitive features of that data instance~\cite{inversion-fred2, attrinf}. 
%
Some attacks are targeted towards stealing information about a deployed model, such as its architecture~\cite{arch1}, trained parameters~\cite{param1-eq} or a general property of the data it was trained on, for instance, if the images used for training were all taken outdoor~\cite{property1}.
%

There is a myriad of methods proposed to tackle these threats. The majority of these methods focus on the data aggregation/dataset publishing and training stages of deep learning. We classify these methods into three classes.  
The first class of methods focuses on sanitizing the data and trying to remove sensitive information from it while maintaining the statistical trends~\cite{kanon, dwork06tcc}. The second class focuses on making the DNN training phase private and protecting the data used for training~\cite{shokriDNN, abadiDNN, pate, hamm, cryptodl, ppmlobf}.
The last class, of which there is only a handful of works, attempts to protect the privacy of the test-time inference phase by protecting the input data (request) that the user sends to a deployed DNN~\cite{osia2, dowlin16icml, shredder}.

In this paper, we first briefly discuss existing attacks and privacy threats against deep learning. Then, we focus on the existing privacy-preserving methods for deep learning and demonstrate that there is a gap in the literature regarding test-time inference privacy. 
There are few other security vulnerabilities which can be exploited in a deep learning model such as adversarial attacks~\cite{adversarialattack}, data poisoning~\cite{datapoisoning}. This work focuses only on privacy specific vulnerability and other such attacks are out of scope of this paper.

%

%
%
%
%
%
\section{Existing Threats}

\begin{figure}
    \centering
    \includegraphics[width=0.5\linewidth]{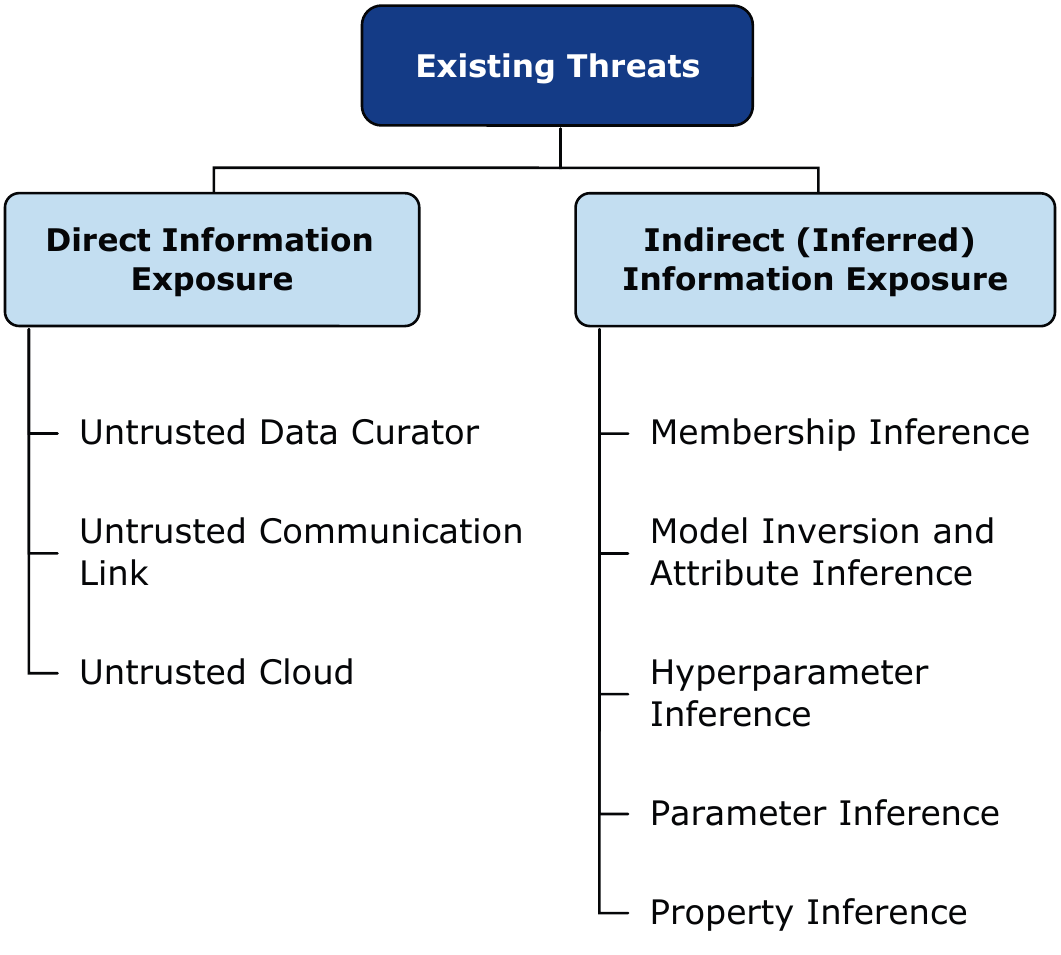}
    \caption{Categorization of existing threats against deep learning}
   \vspace{-4ex}
    \label{fig:threats}
\end{figure}

In this section, we map the space of existing threats against privacy in deep learning and machine learning in general.
While the focus of this survey is privacy-preserving techniques, we provide a brief summary of attacks to better situate the need for privacy protection.
Figure~\ref{fig:threats} shows the landscape of these threats, which we have divided into two main categories of direct and indirect information exposure hazards. 
Direct threats are those where the attacker can gain access to the actual information. In indirect attacks, however, the attacker tries to infer or guess the information and does not have access to the actual information.

\subsection{Direct Information Exposure} 
Direct intentional or unintentional data breaches can occur in many different settings and are not limited to machine learning. 
%
 Dataset breaches through data curators or entities housing the data can be caused unintentionally by hackers, malware, virus, or social engineering, by tricking individuals into handing over sensitive data to adversaries~\cite{breach1}. A study by Intel Security~\cite{intel} demonstrated that employees are responsible for 43\% of data leakage, half of which is believed to be unintentional.
 A malicious party can exploit a system's backdoor to bypass a server’s authentication mechanism and gain direct access to sensitive datasets, or sensitive parameters and models~\cite{Equifax, meltdown, spectre}. The recent hacking of Equifax, for instance, exploited a vulnerability in the Apache Struts software, which was used by Equifax~\cite{Equifax}.   

Data sharing by transmitting conﬁdential
data without proper encryption is an example of data exposure through communication link~\cite{227}. Kaspersky Labs reported in 2018 that they found four million Android apps that were sending unencrypted user profile data to advertisers’ servers~\cite{kaspersky}. 
Private data can also be exposed through the cloud service that receives it to run a process on it, for instance, Machine Learning as a Service (MLaaS). Some of these services do not clarify what happens to the data once the process is finished, nor do they even mention that they are sending the user's data to the cloud, and not processing it locally~\cite{NYTImes}.  

\begin{table*}[]
    \centering
    \footnotesize
    \caption{Properties of some notable attacks against machine learning privacy. MIA denotes Model Inversion Attack in the table.}
    \label{tab:attacks}
     \newcolumntype{L}{>{\centering\arraybackslash}m{0.059\linewidth}} 
  \newcolumntype{O}{>{\centering\arraybackslash}m{0.08\linewidth}} 
  \newcolumntype{D}{>{\centering\arraybackslash}m{0.18\linewidth}} 
  \newcolumntype{R}{>{\arraybackslash}m{0.29\linewidth}}
  \resizebox{\textwidth}{!}{
\begin{tabular}{@{}l@{}l l l l l l l@{}}
	\toprule
	\textbf{Attack} & \textbf{Membership} & \textbf{Model} & \textbf{Hyperparam} & \textbf{Parameter} & \textbf{Property} & \textbf{Access} & \textbf{Access} \\
	& \textbf{Inference} & \textbf{Inversion} & \textbf{Inference} & \textbf{Inference} & \textbf{ Inference} & \textbf{to Model} & \textbf{to Output} \\
    \midrule
		Membership Inference \cite{memberinf} &\CIRCLE & \Circle & \Circle & \Circle & \Circle & Blackbox &	Logits	\\
	Measuring Membership Privacy\cite{memberinf2} &\CIRCLE & \Circle & \Circle & \Circle & \Circle & Blackbox &	Logits	\\
	ML-Leaks\cite{memberinf3} &\CIRCLE & \Circle & \Circle & \Circle & \Circle & Blackbox &	Logits	\\
	The Natural Auditor~\cite{memberinf4} &\CIRCLE & \Circle & \Circle & \Circle & \Circle & Blackbox &	Label	\\
	LOGAN~\cite{memberinf5} &\CIRCLE & \Circle & \Circle & \Circle & \Circle & Both &	Logits	\\
	Data Provenance~\cite{songProv} &\CIRCLE & \Circle & \Circle & \Circle & \Circle & Blackbox &	Logits	\\
	Privacy Risk in ML~\cite{attrinf} &\CIRCLE & \CIRCLE & \Circle & \Circle & \Circle & Whitebox &	Logits+Auxilary	\\
    Fredrikson et al.
	 \cite{pharmacogenetics} &\Circle & \CIRCLE & \Circle & \Circle & \Circle & Blackbox &	Logits	\\
	MIA w/ Confidence Values \cite{inversion-fred2} &\Circle & \CIRCLE & \Circle & \Circle & \Circle & Both &	Logits	\\

	Adversarial NN Inversion~\cite{Yang2019AdversarialNN} &\Circle & \CIRCLE & \Circle & \Circle & \Circle & Blackbox &	Logits	\\
	Updates-Leak~\cite{inversion-salem} &\Circle & \CIRCLE & \Circle & \Circle & \Circle & Blackbox &	Logits	\\
	Collaborative Inference MIA~\cite{collinfMI} &\Circle & \CIRCLE & \Circle & \Circle & \Circle & Both &	Logits	\\

    The Secret Sharer~\cite{property19} &\Circle & \Circle & \Circle & \Circle & \CIRCLE & Blackbox &	Logits	\\
	Property Inference on FCNNs~\cite{property1}  &\Circle & \Circle & \Circle & \Circle & \CIRCLE & Whitebox &	Logits	\\
	Hacking Smart Machines w~\cite{property2} &\Circle & \Circle & \Circle & \Circle & \CIRCLE & Whitebox &	Logits	\\
	 Cache Telepathy~\cite{arch1} &\Circle & \Circle & \Circle & \CIRCLE & \Circle & Blackbox &	Logits	\\
	Stealing Hyperparameters~\cite{stealing19} &\Circle & \Circle & \Circle & \CIRCLE & \Circle & Blackbox &	Logits	\\
	Stealing ML Models~\cite{param1-eq} &\Circle & \Circle & \CIRCLE & \CIRCLE & \Circle & Blackbox &	Label	\\
	\bottomrule
\end{tabular}
}
\end{table*}

\subsection{Indirect (Inferred) Information Exposure} 
As shown in figure~\ref{fig:threats}, we categorize indirect attacks into 5 main groups of membership inference, model inversion, hyperparameter inference, parameter inference, and property inference attacks. Table~\ref{tab:attacks} shows a summary of different attacks and their properties. The ``Access to Model'' column determines whether the attack needs white-box or black-box access to model to successfully mount. White-box access assumes access to the full target model, whereas black-box assumes only query access to the model, without knowledge on the architecture or parameters of the target model. The last column shows whether the attacker needs access to the output confidence values of the model (the probabilities, logits), or whether only the predicted labels suffice.

\subsubsection{Membership Inference}
\label{sec:memberinf}
Given a data instance and (black-box or white-box) access to a pre-trained target model, a membership inference attack speculates whether or not the given data instance has contributed to the training step of the target model.
Shokri et al.~\cite{memberinf} propose the first membership inference attack on machine learning where they consider an attacker who has black-box query access to the target model and can obtain confidence scores (probability vector) for the queried input. The attacker uses this confidence score to deduce the participation of given data in training.
They first train shadow models on a labeled dataset that can be generated using three methods: model inversion attack (we will see next), statistics-based synthesis (through assumptions about the underlying distribution of training set), or noisy real data.  
Using these shadow models, the attacker trains an ``attack model'' that distinguishes the participation of a data instance in the training set of the shadow models. Lastly, for the main inference attack, the attacker makes queries to the target deployed model to receive confidence scores for each given input data instance and infers whether or not the input was part of the target training data. 
This attack is built on the assumption that if a record was used in the training of a model, it would yield a higher confidence score, than a record which was not seen before by the model.

Some studies ~\cite{memor-memb-1, memor-memb-2, attrinf} attribute membership inference attacks to the generalization gap, the over-fitting of the model, and data memorization capabilities of neural networks. Deep neural networks have been shown to memorize the training data~\cite{memorize, casey-mem, mary-mem}, rather than learning the latent properties of the data, which means they often tend to over-fit to the training data. 
%
Long et al. ~\cite{memberinf2} propose an approach which more accurately tests the membership of a given instance. They train the shadow models with and without this given instance, and then at inference time the attacker tests to see if the instance was used for training the target model, similar to Shokri et al.'s approach. More recently, Salem et al.~\cite{memberinf3} propose a more generic attack that could relax the main requirements in previous attacks (such as using multiple shadow models, knowledge of the target model structure, and having a dataset from the same distribution as the target model’s training data), and show that such attacks are also applicable at a lower cost, without significantly degrading their effectiveness.

Membership inference attacks do not always need access to the confidence values (logits) of the target model, as shown by  Song \& Shmatikov in a recent attack~\cite{memberinf4}, which can detect with very few queries to a model if a particular user’s texts were used to train it.

 Yeom et al~\cite{attrinf} suggest a membership inference attack for cases where the attacker can have white-box access to the target model and know the average training loss of the model. In this attack, for an input record, the attacker evaluates the loss of the model and if the loss is smaller than a threshold (the average loss on the training set), the input record is deemed part of the training set. 
Membership inference attacks can also be applied to Generative Adversarial Networks (GANs), as shown by Hayes et al.~\cite{memberinf5}.
%


\begin{figure}
    \centering
    \includegraphics[width=0.8\linewidth]{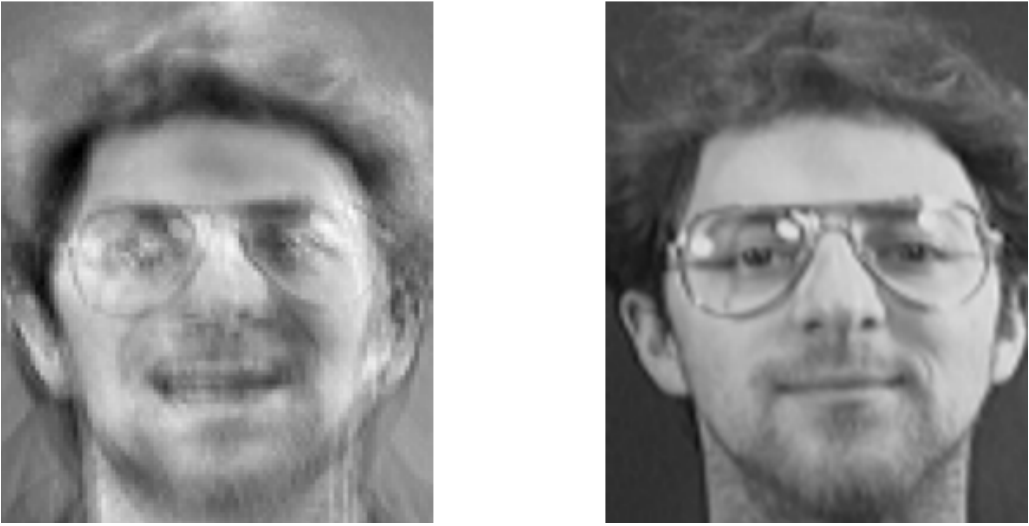}
    \caption{The image on the left was recovered using the model inversion attack of Fredrikson et al.~\cite{inversion-fred2}. The image on the right shows an image from the training set. The attacker is given only the person’s name and access to a facial recognition system
that returns a class confidence score~\cite{inversion-fred2}.}
    \vspace{-3ex}
    \label{fig:inversion}
\end{figure}

\subsubsection{Model Inversion and Attribute Inference}

Model inversion and attribute inference attacks are against attribute privacy, where an adversary tries to infer sensitive attributes of given data instances from a released model and the instance's non-sensitive attributes~\cite{inversion-meth}. 
The most prominent of these attacks is against a publicly-released linear regression model, where Fredrikson et al.~\cite{pharmacogenetics} invert the model of a medicine (Warfarin) dosage prediction task.
They recover genomic information about the patient, based on the model output and several other non-sensitive attributes (e.g., height, age, weight). This attack can be applied with only black-box API access to the target model.
Fredrikson et al. formalize this attack as maximizing the posterior probability estimate of the sensitive attribute. In other words, the attacker assumes that features $f_1$ to $f_{d-1}$, of the $f_d$ features of each data instance are non-sensitive. the attacker then tries to maximize the posterior probability of feature $f_d$, given the nonsensitive features of $f_1$ to $f_{d-1}$, and the model output.

In another work, given white-box access to a neural network, Fredrikson et al.~\cite{inversion-fred2} show that they could extract instances of training data, from observed model predictions. Figure~\ref{fig:inversion} shows a recovered face image that is similar to the input image and was reconstructed by utilizing the confidence score of the target model.
Yeom et al.~\cite{attrinf} also propose an attribute inference attack, built upon the same principle used for their membership inference attack, mentioned in Section~\ref{sec:memberinf}. The attacker evaluates the model’s loss on the input instance for different values of the sensitive attribute and infers the value that yields a loss value similar to that outputted by the original data, as the sensitive value.
Salem et al.~\cite{inversion-salem} suggest a model inversion attack on online-learning, using a generative adversarial network and based on the difference between a model, before and after a gradient update.
In the same direction, Brockschmidt et al.~\cite{brockschmidt2020analyzing} demonstrate the information leakage of updates to language models.
More recently, He et al.~\cite{collinfMI} propose a new set of attacks to compromise the privacy of test-time inference queries, in collaborative deep learning systems where a DNN is split and distributed to different participants.
This scheme is called split learning~\cite{main-split}, which is discussed in Section~\ref{sec:models}. They demonstrate that with their attack, one malicious participant can recover an arbitrary input fed into this system, even with no access to other participants’ data or computations.

\subsubsection{Model Stealing: Hyperparameter and Parameter Inference}

Trained models are considered intellectual properties of their owners and can be considered confidential in many cases~\cite{financial}, therefore extracting the model can be considered a privacy breach. Apart from this, as discussed earlier, DNNs are shown to memorize information about their training data, therefore exposing the model parameters could lead to exposure of training data. A model stealing attack is meant to recover the model parameters via black-box access to the target model.
Tramer et al.~\cite{param1-eq} devise an attack that finds parameters of a model given the observation of its predictions (confidence values). Their attack tries to find parameters of the model through equation solving, based on pairs of input-outputs. This attack cannot be mounted in a setting where the confidence values are not provided.
Hyperparameter stealing attacks try to find the hyperparameters
used during the model training, such as the regularization coefficient ~\cite{stealing19} or model architecture~\cite{arch1}. 

\subsubsection{Property Inference} 

This class of attacks tries to infer specific patterns of information from the target model. An example of these attacks is the memorization attack that aims to find sensitive patterns in the training data of a target model~\cite{property19}.  These attacks have been mounted on Hidden Markov Models (HMM)
and Support Vector Machines (SVM) ~\cite{property2} and neural networks~\cite{property1}.


\section{Privacy-Preserving Mechanisms}

\begin{figure}
    \centering
    \includegraphics[width=0.8\linewidth]{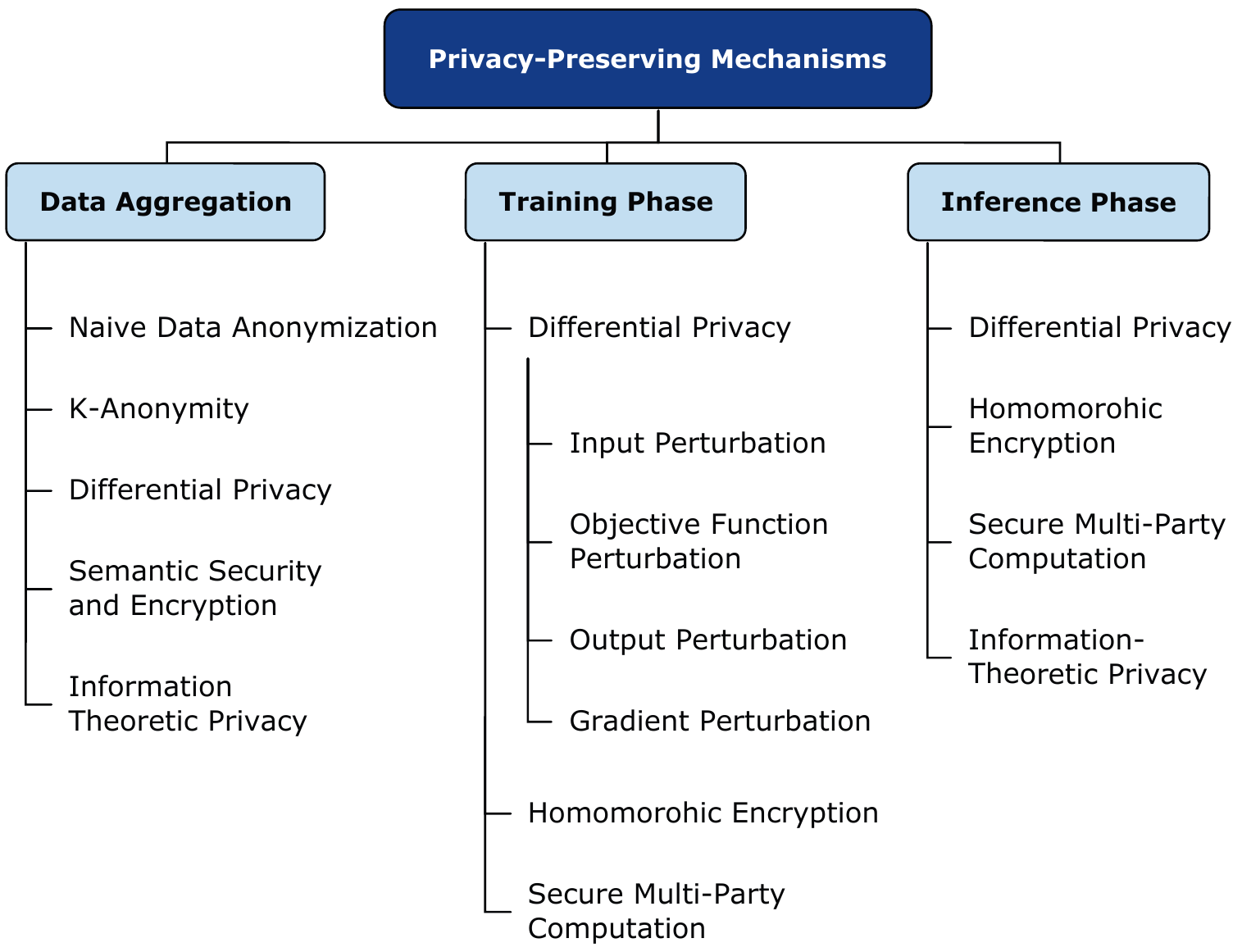}
    \caption{Categorization of privacy-preserving schemes for deep learning.}
   \vspace{-2ex}
    \label{fig:training-hir}
\end{figure}

In this section, we review the literature of privacy-preserving mechanisms for deep learning and machine learning in general. Figure~\ref{fig:training-hir} shows our classification of the landscape of this field. We divide the literature into three main groups. The first is private data aggregation methods, which aim at collecting data and forming datasets, while preserving the privacy of the contributors~\cite{kanon, dwork06tcc}. 
The second group, which is comprised of a large body of work focuses on devising mechanisms that make the training process of models private so that sensitive information about the participants of the training dataset would not be exposed.
Finally, the last group aims at the test-time inference phase of deep learning. It tries to protect the privacy of users of deployed models, who send their data to a trained model for having a given inference service carried out. 

\subsection{Data Aggregation}\label{sec:data}

Here, we introduce the most prominent data privacy-preserving mechanisms. Not all these methods are applied to deep learning, but we briefly discuss them for the sake of comprehensiveness. These methods can be broadly divided into two groups of context-free privacy and context-aware. Context-free privacy solutions, such as differential privacy, are unaware of the specific context or the purpose that the data will be used for. Whereas context-aware privacy solutions, such as information-theoretic privacy, are aware of the context where the data is going to be used, and can
achieve an improved privacy-utility tradeoff~\cite{lalitha-pv}.


\subsubsection{Naive Data Anonymization}
What we mean by naive anonymization in this survey is the removal of identifiers from data, such as the names, addresses, and full postcodes of the participants, to protect privacy. This method was used for protecting patients while processing medical data and has been shown to fail on many occasions~\cite{netflix, kanon, homer2008resolving}. 
Perhaps the most prominent failure is the Netflix prize case, where Narayanan \& Shmatikov apply their de-anonymization technique to the Netflix Prize dataset. 
This dataset contains anonymous movie ratings of 500,000 subscribers of Netflix. They showed that an adversary with auxiliary knowledge (from the publicly available Internet Movie Database records)   about individual subscribers can easily identify the user and uncover potentially sensitive information~\cite{netflix}.

\subsubsection{K-Anonymity}

A dataset has k-anonymity property if each participant's information cannot be distinguished from at least $k-1$ other participants whose information is in the dataset~\cite{kanon}. 
K-anonymity means that for any given combination of attributes that are available to the adversary (these attributes are called quasi-identifiers), there are at least k rows with the exact same set of attributes.
K-anonymity has the objective of impeding re-identification. However, k-anonymization has been shown to perform poorly on the anonymization of high-dimensional datasets~\cite{kanon-fail}. This has led to privacy notions such as l-diversity~\cite{ldiver} and t-closeness~\cite{tclose}, which are out of the scope of this survey.

\subsubsection{Differential Privacy}\label{sec:dp}
\begin{defn}\textbf{$\epsilon$-Differential Privacy ($\epsilon$-DP).} For $\epsilon\geq 0$, an algorithm $A$ satisfies $\epsilon$-DP~\cite{dwork06euro,dwork06tcc} if and only if for any pair of datasets $D$ and $D'$ that differ in only one element:
\begin{equation}\label{eq:dp}
\mathcal{P}[A(D)=t] \leq e^\epsilon \mathcal{P}[A(D')=t]\;\;\;\forall t
\end{equation}
where, $\mathcal{P}[A(D)=t]$ denotes the probability that the algorithm $A$ outputs $t$. In this setup, the quantity $ln\frac{\mathcal{P}[A(D)=t]}{\mathcal{P}[A(D')=t]}$ is named the privacy loss. DP tries to approximate the effect of an individual opting out of contributing to the dataset, by ensuring that any effect due to the inclusion of one’s data is small. One of the widely used DP mechanisms when dealing with numerical data is the Laplace mechanism.\end{defn}

\begin{defn}\textbf{Laplace Mechanism.}~\cite{dwork06tcc} Given a target function $f$ and a fixed $\epsilon\geq 0$, the randomizing algorithm $A_f(D)=f(D)+x$ where $x$ is a perturbation random variable drawn from a Laplace distribution $Lap(\mu,\frac{\Delta_f}{\epsilon})$, is called the Laplace Mechanism and is $\epsilon$-DP. Here, $\Delta_f$ is the global \textbf{sensitivity} of function $f$, and is defined as $\Delta_f=\sup |f(D)-f(D')|$ over all the dataset pairs $(D,D')$ that differ in only one element. Finding this sensitivity is not always trivial, specifically if the function $f$  is a deep neural network, or even a number of layers of it~\cite{pixeldp}.\end{defn}

Differential privacy satisfies a composition property that states when two mechanisms with privacy budgets $\epsilon_1$ and $\epsilon_2$ are applied to the same datasets, together they use a privacy budget of $\epsilon_1$ + $\epsilon_2$. 
As such, composing multiple differentially private mechanisms consumes a linearly increasing privacy budget.
It has been shown that tighter privacy bound for composition can be reached, so that the privacy budget decreases sub-linearly~\cite{strong1, strong2}. There are multiple variants of the conventional $\epsilon$-differential privacy which have been proposed to provide a tighter analysis of the privacy budget under composition. One of them is differential privacy with Advanced Composition (AC)~\cite{dwork14book}, which allows an additive leakage probability parameter $\delta$ to the right-hand side of Equation~\ref{eq:dp}.
%

Differential privacy can also be achieved without the need to trust a centralized server by having each participant apply a differentially private randomization to their data themselves, before sharing it.
This model is named the local model of differential privacy, and the method ``randomized response'' is shown to be locally differentially private~\cite{randomize}. Local differential privacy has been deployed on many systems for gathering statistics privately~\cite{rappor, joseph}. For instance, Google uses a technique named RAPPOR \cite{rappor} to allow web browser developers to privately collect usage statistics.
A large body of Differentially private mechanisms has been proposed for various applications~\cite{Gopi2020DifferentiallyPS, abadiDNN, dprnn}. Triastcyn \& Faltings present a technique that generates synthetic datasets that still have statistical properties of the real data while providing differential privacy guarantees with respect to this data~\cite{dpdataset}. 

A generalized version of differential privacy called Pufferfish was proposed by~\cite{puffermain}. The Pufferfish framework can be used to create new privacy definitions tailored for specific applications~\cite{puffer2}, such as Census data release.
Another framework that is also an adaptation of differential privacy for location obfuscation is dubbed geo-indistinguishablity~\cite{geo-indist}.
Geo-indistinguishablity relaxes differential privacy's guarantees by introducing a distance metric, $d$, which is multiplied by $\epsilon$ in Equation~\ref{eq:dp}. This distance permits adjusting the obfuscation such that the probability of being obfuscated to a closer point (being mapped to a point closer to the real value) is higher than being obfuscated to a far point.

\subsubsection{Semantic Security and Encryption}\label{sec:semantic}
Semantic security~\cite{semantic} (computationally secure) is a standard privacy requirement of encryption schemes which states that the advantage (a measure of how successfully an adversary can attack a cryptographic algorithm) of an adversary with background information should be cryptographically small. Semantic security is theoretically possible to break but it is infeasible to do so by any known practical means~\cite{pvpv}. Secure Multiparty Computation (SMC), which we discuss in Section~\ref{sec:train}, is based on semantic security definition~\cite{emory}.

\subsubsection{Information-Theoretic Privacy}\label{sec:it}
Information-theoretic privacy is a context-aware privacy solution. Context-aware solutions explicitly model the dataset statistics, unlike context-free solutions that assume worst-case dataset statistics and adversaries. There is a body of work studying information-theoretic based methods for both privacy and fairness, where privacy and fairness are provided through information degradation, through obfuscation or adversarial learning and demonstrated by mutual information reduction ~\cite{lalitha-measure, lalitha-datagen, lalitha-pv, lalitha-fairness, smart-meter, ObfuscationVI,  DeepObfuscatorAT, Mirjalili2019FlowSANPS, Roy2019MitigatingIL, Wu2018TowardsPV, ganobfmi}. 
Huang et al. introduce a context-aware privacy framework
called generative adversarial privacy (GAP), which leverages generative adversarial networks (GANs) to generate privatized datasets. Their scheme comprises of a sanitizer that tries to remove private attributes, and an adversary that tries to infer them~\cite{lalitha-pv}. They show that the privacy mechanisms learned from data (in a generative adversarial fashion) match the theoretically optimal ones.
%

\subsection{Training Phase}\label{sec:train}
\begin{table*}[]
    \centering
    \fontsize{6}{6}
    \caption{Categorization of some notable privacy-preserving mechanisms for training. In the table, the following abbreviations have been used:  ERM for Empirical Risk Minimization, GM for Generative Model, AE for Auto Encoder, LIR for Linear Regression, LOR for Logistic Regression, LM for Linear Means, FLD  for Fisher’s Linear Discriminant, NB for Naive Bayes and RF for Random Forest.}
    \label{tab:training}
     \newcolumntype{L}{>{\centering\arraybackslash}m{0.017\linewidth}} 
  \newcolumntype{D}{>{\arraybackslash}m{0.32\linewidth}} 
\newcolumntype{K}{>{\arraybackslash}m{0.27\linewidth}} 
\newcolumntype{P}{>{\arraybackslash}m{0.27\linewidth}}
\footnotesize
\begin{tabular}{@{}D@{}LLLP K@{}}
	\toprule
	\textbf{Method} & \textbf{DP} & \textbf{SMC} & \textbf{HE} & \textbf{Dataset(s)} & \textbf{Task}\\
    \midrule
	 DPSGD~\cite{abadiDNN} &\CIRCLE & \Circle & \Circle	& MNIST, CIFAR-10 & Image Classification w/ DNN \\
	DP LSTM~\cite{DPLSTM} &\CIRCLE & \Circle & \Circle & Reddit Posts &  Language Model w/ LSTMs\\
	DP LOR~\cite{chaudhuriLogistic} &\CIRCLE & \Circle & \Circle & Artificial Data	& Logistic Regression\\
	DP ERM~\cite{chaudhuriERM} &\CIRCLE & \Circle & \Circle &	Adult, KDD-99 & Classification w/ ERM\\
	DP GAN~\cite{dpgan} &\CIRCLE & \Circle & \Circle	& MNIST, MIMIC-III & Data Generation w/ GAN \\
	DP GM~\cite{dpgm} &\CIRCLE & \Circle & \Circle & MNIST, CDR, TRANSIT & Data Generation w/ GM \\
	DP AE~\cite{dpa} &\CIRCLE & \Circle & \Circle	 & Health Social Network Data & Behaviour Prediction w/ AE \\
	DP Belief Network\cite{dpbn} &\CIRCLE & \Circle & \Circle	& YesiWell, MNIST&   Classification w/ DNN\\
	Adaptive Laplace Mechanism\cite{adlm} &\CIRCLE & \Circle & \Circle	& MNIST, CIFAR-10&  Image Classification w/ DNN\\
	PATE~\cite{pate} &\CIRCLE & \Circle & \Circle	 & MNIST, SVHN & Image Classification w/ DNN\\
	Scalable Learning w/ PATE~\cite{pate2} &\CIRCLE & \Circle & \Circle	& MNIST, SVHN, Adult, Glyph & Image Classification w/ DNN\\
	DP Ensemble~\cite{hamm} &\CIRCLE & \Circle & \Circle	 &KDD-99, UCI-HAR, URLs & Classification w/ ERM\\
	SecProbe~\cite{colab} &\CIRCLE & \Circle & \Circle	 & US, MNIST, SVHN & Regress. \& Class. w/ DNN\\
	Distributed DP~\cite{clinical} &\CIRCLE & \Circle & \Circle	 & eICU, TCGA&  Classification w/ DNN\\
	DP model publishing~\cite{dppublish} &\CIRCLE & \Circle & \Circle	 & MNIST, CIFAR & Image Classification w/ DNN \\
	DP federated learning~\cite{dpfd} &\CIRCLE & \Circle & \Circle	 & MNIST & Image Classification w/ DNN \\
	ScalarDP, PrivUnit~\cite{reconfd} &\CIRCLE & \Circle & \Circle	 & MNIST, CIFAR & Image Classification w/ DNN \\
	 DSSGD~\cite{shokriDNN} &\CIRCLE & \Circle & \Circle	 & MNIST, SVHN & Image Classification w/ DNN \\
	Private Collaborative NN~\cite{pvcolab-dpsmc} &\CIRCLE & \CIRCLE & \Circle	 & MNIST & Image Classification w/ DNN  \\ 
	Secure Aggregation for ML~\cite{smc-fed} &\Circle & \CIRCLE & \Circle	 & - & Federated Learning \\
	QUOTIENT~\cite{qou} &\Circle & \CIRCLE & \Circle	 & MNIST, Thyroid,  Credit & Classification w/ DNN\\
	SecureNN~\cite{secureNN} &\Circle & \CIRCLE & \Circle	 &  MNIST & Image Classification w/ DNN\\
	ABY3~\cite{ABY} &\Circle & \CIRCLE & \Circle	 & MNIST  & LIR, LOR, NN \\
	Trident~\cite{Rachuri2019TridentE4} &\Circle & \CIRCLE & \Circle	 & MNIST, Boston Housing & LIR, LOR, NN \\
	SecureML~\cite{secureml} &\Circle & \CIRCLE & \CIRCLE	& MNIST, Gisette, Arcene & LIR, LOR, NN \\
	Deep Learning w/ AHE~\cite{dlhe} &\Circle & \Circle & \CIRCLE & MNIST& Image Classification w/ DNN \\
	ML Confidential~\cite{mlconf1} &\Circle & \Circle & \CIRCLE & Wisconsin Breast Cancer & LM, FLD \\
	Encrypted Statistical ML~\cite{mlconf2} &\Circle & \Circle & \CIRCLE & 20 datasets from UCI ML & LOR, NB, RF\\
	CryptoDL~\cite{cryptodl} &\Circle & \Circle & \CIRCLE	& MNIST, CIFAR-10 & Image Classification w/ DNN\\
	DPHE~\cite{HEvisual} &\Circle & \Circle & \CIRCLE	& Caltech101/256, CelebA & Image Classification w/ SVM\\

	\bottomrule
\end{tabular}

\end{table*}
The literature surrounding private training of deep learning, and machine learning can be categorized based on the guarantee that these methods provide, which is most commonly either based on differential privacy (Section~\ref{sec:dp}) or semantic security and encryption (Section~\ref{sec:semantic}).
Privacy using encryption is achieved by doing computation over encrypted data. The two most common methods for this are Homomorphic Encryption (HE) and Secure Multi-Party Computation (SMC). 
\textbf{Homomorphic Encryption (HE).}  HE~\cite{fullhomomorphic} allows computation over encrypted data. A client can send their data, in an encrypted format, to a server and the server can compute over this data without decrypting it, and then send a ciphertext (encrypted result) to the client for decryption. HE is extremely compute-intensive and is therefore not yet deployed in many production systems~\cite{xonn, epic}. 

\textbf{Secure Multi-Party Computation (SMC).}  SMC attempts at designing a network of computing parties (not all of which the user necessarily has to trust) that carry out a given computation and makes sure no data leaks. Each party in this network has access to only an encrypted part of the data. SMC ensures that as long as the owner of the data trusts at least one of the computing systems in the network, their input data remain secret.
Simple functions can easily be computed using this scheme. Arbitrarily complex function computations can also be supported, but with an often prohibitive computational cost~\cite{epic}.

In this survey, we divided the literature of private training into three groups of methods that employ: 1) Differential Privacy (DP),  2) Homomorphic Encryption (HE) and 3) Secure Multi-Party Computation (SMC). Table~\ref{tab:training} shows this categorization for the literature we discuss in this section.

\subsubsection{Differential Privacy}
\begin{figure}
    \centering
    \includegraphics[width=\linewidth]{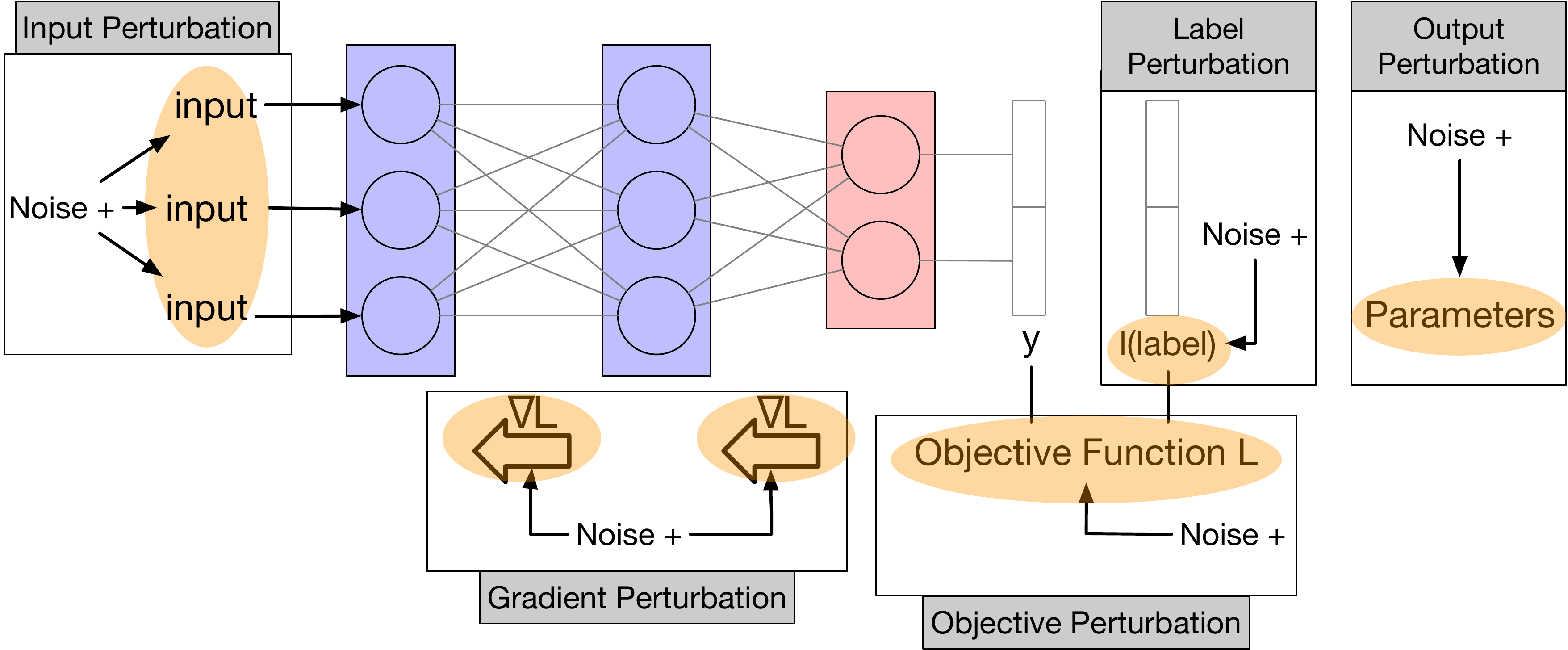}
    \caption{ Overview of how a deep learning framework works and how differential privacy can be applied to different parts of the pipeline.}
    \label{fig:flow}
\end{figure}
This section briefly discusses methods for modifying deep learning algorithms to satisfy differential privacy. Figure~\ref{fig:flow} shows an overview of a deep learning framework. As can be seen, the randomization required for differential privacy (or the privacy-preserving noise) can be inserted in five places: to the input, to the loss/objective function, to the gradient updates, to the output (the optimized parameters of the trained model) and to the labels~\cite{pate}. 

\textbf{Input perturbations} can be considered equivalent to using a sanitized dataset (discussed in Section~\ref{sec:data}) for training. 
\textbf{objective function perturbation} and \textbf{output perturbation} are explored for machine learning tasks with convex objective functions. For instance in the case of logistic regression, Chaudhuri et al. prove that objective perturbation requires sampling noise in the scale of $\frac{2}{n\epsilon}$, and output perturbation requires sampling noise in the scale of $\frac{2}{n\lambda\epsilon}$, where $n$ is the number of samples and $\lambda$ is the regularization coefficient~\cite{chaudhuriERM}. More recently, Iyengar et al~\cite{iyengar2019towards} propose a more practical and general objective perturbation approach, and benchmark it using high-dimensional real-world data. 
In deep learning tasks, due to the non-convexity of the objective function, calculating the sensitivity of the function (which is needed to determine the intensity of the added noise) becomes non-trivial.
One solution is replacing the non-convex function with an approximate convex polynomial function ~\cite{dpa, dpbn, adlm} and then using objective function perturbation. This approximation limits the capabilities and the utility that a conventional DNN would have. 
Given discussed limitations, \textbf{gradient perturbation} is the approach that is widely used for private training in deep learning. 
Applying perturbations on the gradients requires the gradient norms to be bounded, and since in deep learning tasks the gradient could be unbounded, clipping is usually used to alleviate this issue.  
%

Shokri et al. showed that deep neural networks can be trained in a distributed manner and with perturbed parameters to achieve privacy~\cite{shokriDNN}, but their implementation requires $\epsilon$ proportional to the size of the target model, which can be in the order of couple millions.
Abadi et al.~\cite{abadiDNN} propose a mechanism dubbed the ``moments accountant (MA)'', for bounding the cumulative privacy budget of sequentially applied differentially private algorithms, over deep neural networks. 
The moments accountant uses the moment generating function of the privacy loss random variable to keep track of a bound on the privacy loss during composition. MA operates in three steps: first, it calculates the moment generating functions for the algorithms $A_1$, $A_2$,.., which are the randomizing algorithms. 
It then composes the moments together through a composition theorem, and finally, finds the best leakage parameter ($\delta$) for a given privacy budget of $\epsilon$.
The moments accountant is widely used in different DP mechanisms for private deep learning.  Papernot et al. use MA to aid bounding the privacy budget for their teacher ensemble method that uses noisy voting and \textbf{label perturbation}~\cite{pate,pate2}. MA is also employed by the works~\cite{clinical, dppublish, dpfd, dpgan, dpgm, reconfd} all of which use perturbed gradients. 

More recently, Bu et al. apply the Gaussian Differential Privacy (GDP) notion introduced by Dong et al.~\cite{Dong2019GaussianDP} to deep learning~\cite{Bu2019DeepLW} to achieve a more refined analysis of neural network training, compared to that of Abadi et al.~\cite{abadiDNN}.
They analyze the privacy budget exhaustion of private DNN training using Adam optimizer, without the need of developing sophisticated techniques such as the moments accountant. They demonstrate that GDP allows for a new privacy analysis that improves on the moments accountant analysis and provides better guarantees (i.e. lower $\epsilon$ values). 

Inherently, applying differential privacy to deep learning yields loss of utility due to the addition of noise and clipping. Bagdasaryan et al. have demonstrated that this loss in utility is disparate across different sub-groups of the population, with different sizes~\cite{bagdasaryan2019differential}. 
They experimentally show that sub-groups with less training samples (less representation) lose more accuracy, compared to well-represented groups, i.e. the poor get poorer. 

There is a body of work that tries to experimentally measure and audit the privacy brought by differentially private learning algorithms~\cite{Jagielski2020AuditingDP, jayaraman2019evaluating}. Jagielski et al.~\cite{Jagielski2020AuditingDP} investigate whether DP-SGD offers better privacy in practice than what is guaranteed by its  analysis, using data poisoning attacks. 
Jayaraman et al~\cite{jayaraman2019evaluating} apply membership and attribute inference attacks on multiple differentially private machine learning and deep learning algorithms, and compare their performance. 

\subsubsection{Homomorphic Encryption}\label{sec:het}
There are only a handful of works that exploit solely homomorphic encryption for private training of machine learning models~\cite{mlconf1, mlconf2, cryptodl}. Graepel et al. use a Somewhat HE (SHE) scheme to train  Linear Means (LM) and Fisher’s Linear Discriminate (FLD) classifiers~\cite{mlconf1}. HE algorithms have some limitations in terms of the functions they can compute (for instance they cannot implement non-linearities). For that reason,  Graepel et al. propose division-free algorithms and focus on simple classifiers and not complex algorithms such as neural networks. 

Hesamifard et al.~\cite{cryptodl} try to exploit HE for deep learning tasks. They introduce methods for approximating the most commonly used neural network activation functions (ReLU, Sigmoid, and Tanh) with low degree polynomials. This is a crucial step for designing efficient homomorphic encryption schemes. They then train convolutional neural networks with those approximate polynomial functions and finally, implement convolutional neural networks over encrypted data and measure the performance of the models.
%

\subsubsection{Secure Multi-Party Computation (SMC)}

A trend in research on private and secure computation consists of designing custom protocols for applications such as linear and logistic regression ~\cite{secureml} and neural network training and inference~\cite{secureml, qou, Shamsabadi2020PrivEdgeFL}. 
These methods usually target settings where different datasets from different places are set to train a model together, or where computation is off-loaded to a group of computing servers that do not collude with each other. SMC requires that all participants be online at all times, which requires a significant amount of communication~\cite{fdsurvey}.
Mohassel \& Zhang proposed SecureML which is a privacy-preserving stochastic gradient descent-based method to privately train machine learning algorithms such as linear regression, logistic regression and neural networks in multiparty computation settings. SecureML uses secret sharing to achieve privacy during training.  
In a more recent work~\cite{ABY}, Mohassel et al design protocols for secure three-party training of DNNs with a majority of honest parties. 
Agrawal et al. propose QUOTIENT~\cite{qou} where their goal is to design an optimization algorithm alongside a secure computation protocol customized for it, instead of a conventional approach which is using encryption on top of existing optimization algorithms.

\subsection{Inference Phase}
\begin{table*}[]
    \centering
    \footnotesize
    \fontsize{8}{8}
    \caption{Categorization of some notable privacy-preserving mechanisms for inference. In this table, NB is short for Naive Bayes, and DT is short for Decision Tree.}
    \label{tab:inference}
     \newcolumntype{L}{>{\centering\arraybackslash}m{0.013\linewidth}} 
  \newcolumntype{D}{>{\arraybackslash}m{0.31\linewidth}} 
  
  \newcolumntype{P}{>{\arraybackslash}m{0.26\linewidth}} 
 
  \newcolumntype{K}{>{\arraybackslash}m{0.27\linewidth}} 
\resizebox{\textwidth}{!}{  
\begin{tabular}{@{}l@{}l@{}l@{}l@{}l@{}l@{}l@{}}
	\toprule
	\textbf{Method} & \textbf{DP } & \textbf{SMC } & \textbf{HE } & \textbf{IT } & \textbf{Dataset(s)} & \textbf{Task}\\
    \midrule
	ARDEN~\cite{notjust} & \CIRCLE & \Circle & \Circle& \Circle	& MNIST, CIFAR-10, SVHN & Image  Classification w/ DNN \\
	Cryptonets~\cite{cryptonets} &\Circle & \Circle & \CIRCLE& \Circle	& MNIST &  Image Classification w/ DNN\\ 
	Private Classification~\cite{pvclass} &\Circle & \Circle & \CIRCLE	& \Circle & MNIST &  Image Classification w/ DNN\\
	TAPAS~\cite{TAPAS} &\Circle & \Circle & \CIRCLE	& \Circle&MNIST, Faces, Cancer, Diabetes &  Image Classification w/ DNN\\
	FHE–DiNN~\cite{fhdnn} &\Circle & \Circle & \CIRCLE	& \Circle& MNIST & Image Classification w/ DNN\\
	Face Match~\cite{Facematch} &\Circle & \Circle & \CIRCLE	& \Circle& LFW, IJB-A, IJB-B, CASIA & Face recognition with CNNs\\
	Cheetah~\cite{reagen2020cheetah} &\Circle & \Circle & \CIRCLE	& \Circle& MNIST, Imagenet& Image Classification w/ DNN\\	
    EPIC~\cite{epic} &\Circle & \CIRCLE & \Circle	& \Circle& CIFAR-10, MIT, Caltech& Image Classification w/ DNN\\
	DeepSecure~\cite{deepsecure} &\Circle & \CIRCLE & \Circle&	\Circle&  MNIST, UCI-HAR&  Classification w/ DNN\\
    XONN~\cite{xonn} &\Circle & \CIRCLE & \Circle&	\Circle& MNIST, CIFAR-10 & Image Classification w/ DNN  \\
	Chameleon~\cite{chamel} &\Circle & \CIRCLE &\Circle 	& \Circle& MNIST, Credit Approval&  Classification w/ DNN and SVM\\
	CRYPTFLOW~\cite{Kumar2020CrypTFlowST} &\Circle & \CIRCLE &\Circle 	& \Circle& MNIST,CIFAR, ImageNet&  Classification w/ DNN\\
	Classification over Encrypted Data\cite{ovenc} &\Circle & \CIRCLE & \CIRCLE	& \Circle& Wisconsin Breast Cancer & Classification w/  NB, DT \\
	 MiniONN~\cite{minionn} &\Circle & \CIRCLE & \CIRCLE	& \Circle& MNIST, CIFAR-10 &  Image  Classification w/ DNN  \\
	GAZELLE~\cite{gazelle} &\Circle & \CIRCLE & \CIRCLE	& \Circle& MNIST, CIFAR-10& Image  Classification w/ DNN \\
	DELPHI~\cite{delphim} &\Circle & \CIRCLE & \CIRCLE	& \Circle& CIFAR-10, CIFAR-100& Image  Classification w/ DNN \\
	Shredder~\cite{shredder} &\Circle & \Circle & \Circle&	\CIRCLE& SVHN, VGG-Face, ImageNet &   Classification w/ DNN \\
	Sensor Data Obfuscation~\cite{10.1145/3302505.3310068} & \Circle & \Circle & \Circle& \CIRCLE	& Iphone 6s Accelerometer Data & Activity Recognition w/ DNN\\
	Olympus~\cite{raval2019olympus} & \Circle & \Circle & \Circle& \CIRCLE	& Driving images& Activity Recognition w/ DNN\\
	DPFE~\cite{osia2} &\Circle & \Circle & \Circle&	\CIRCLE& CelebA& Image  Classification w/ DNN \\
	Cloak~\cite{cloak} & \Circle & \Circle & \Circle& \CIRCLE	& CIFAR-100, CelebA, UTKFace & Image  Classification w/ DNN\\

	\bottomrule
\end{tabular}
}

\end{table*}

As shown in Table~\ref{tab:inference} there are fewer works in the field of inference privacy, compared to training. Inference privacy targets systems that are deployed to offer Inference-as-a-Service. In these cases, the deployed system is assumed to be trained and is not to learn anything new from the data provided by the user. It is only supposed to carry out its designated inference task. 
The categorization of literature for inference privacy is similar to training, except that there is one extra group here, named Information-Theoretic (IT) privacy. The works in this group usually offer information-theoretic mathematical or empirical evidence of how their methods operate and help privacy. %
These works are based on the context-aware privacy definition of Section~\ref{sec:it}, and they aim at decreasing the information content in the data sent to the service provider for inference so that there is only as much information in the input as needed for the service and not more. 

One notable difference between training and inference privacy is the difference in the amount of literature on different categories. There seems to be a trend of using differential privacy for training, and encryption methods (HE and SMC) for inference. One underlying reason could be computational complexity and implementation. 
Encryption methods, specifically homomorphic encryption, are shown to be at least two orders of magnitude slower than conventional execution~\cite{gazelle}. That's why adopting them for training will increase training time significantly. Also, as mentioned in Section~\ref{sec:het}, due to approximating non-linear functions, the capabilities of neural networks in terms of performance become limited during training on encrypted data. 
For inference, however, adopting encryption is more trivial, since the model is already trained. Employing differential privacy, and noise addition, however, is less trivial for inference, since it could damage the accuracy of the trained model, if not done meticulously. Below we delve deeper into the literature of each category.
\subsubsection{Differential Privacy}\label{sec:train:dp}

There are very few works using differential privacy for inference. The main reason is that differential privacy offers a worst-case guarantee which requires high-intensity noise (noise with high standard deviation) to be applied to all the segments of the input. This inherently causes performance degradation on pre-trained networks. Wang et al.~\cite{notjust} propose Arden, a data nullification and differentially private noise injection mechanism for inference. 
Arden partitions the DNN across edge device and the cloud. A simple data transformation is performed on the mobile device, while the computation heavy and complex inference relies on the cloud data center. 
Arden uses data nullification, and noise injection to make different queries indistinguishable so that the privacy of the clients is preserved. The proposed scheme requires noisy retraining of the entire network, with noise injected at different layers. 
Since it is complicated to calculate the global sensitivity at each layer of the neural network, the input to the noise injection layer is clipped to the largest possible value created by a member of the training set, on the trained network. 
%


\subsubsection{Homomorphic Encryption} 
CryptoNets is one of the first works in HE inference~\cite{cryptonets}. Dowlin et al. present a method for converting a trained neural network into an encrypted one, named a CryptoNet. This allows the clients of an inference service to send their data in an encrypted format and receive the result, without their data being decrypted. 
CryptoNets allows the use of SIMD (Single Instruction Multiple Data) operations, which increase the throughput of the deployed system. However, for single queries, the latency of this scheme is still high. 

Chabanne et al.~\cite{pvclass} approximate the ReLu non-linear activation function using low-degree polynomials and provide a normalization layer before the activation function, which offers high accuracy. However, they do not show results on the latency of their method. 
More recently, Juvekar et al. propose GAZELLE~\cite{gazelle}, a system with lower latency (compared to prior work) for secure and private neural network inference. GAZELLE combines homomorphic encryption with traditional two-party computation techniques (such as garbled circuits). With the help of its homomorphic linear algebra kernels, which map neural network operations to optimized homomorphic matrix-vector multiplication and convolutions, GAZELLE is shown to be three orders of magnitude faster than CryptoNets. 
Sanyal et al. leverage binarized neural networks to speed-up their HE inference method. They claim that unlike CryptoNets which only protects the data, their proposed scheme can protect the privacy of the model as well.

\subsubsection{Secure Multi-Party Computation (SMC)}

Liu et al. propose MiniONN~\cite{minionn}, which uses additively homomorphic encryption (AHE) in a preprocessing step, unlike GAZELLE which uses  AHE to speed up linear algebra directly.  MiniONN demonstrates a significant performance improvement compared to CryptoNets, without loss of accuracy. However, it is only a two-party computation scheme and does not support computation over multiple parties.
Riazi et al. introduce Chameleon, a two-party computation framework whose vector dot product of signed fixed-point numbers improves the efficiency of prediction in classification methods based upon heavy matrix multiplications. Chameleon achieves a 4.2$\times$ latency improvement over MiniONN.
Most of the efforts in the field of SMC for deep learning are focused on speeding up the computation, as demonstrated above, and also by ~\cite{xonn}, ~\cite{epic},   ~\cite{deepsecure}. The accuracy loss of the aforementioned methods, compared to their pre-trained models is negligible (less than 1\%).

\subsubsection{Information Theoretic Privacy}

Privacy-preserving schemes that rely on information-theoretic approaches usually assume a non-sensitive task, the task that the service is supposed to execute and try to degrade any excessive information in the input data that is not needed for the main inference task~\cite{10.1145/3302505.3310068, malekzadeh, raval2019olympus, osia2, shredder, malekzadeh2020privacy}.  
~\cite{10.1145/3302505.3310068, raval2019olympus, malekzadeh} propose anynymization schemes for protecting temporal sensory data through obfuscation.
Malekzadeh et al.  \cite{10.1145/3302505.3310068} propose a
multi-objective loss function for training deep autoencoders to extract and obfuscate user identity-related information, while preserving the utility of the sensor data. The training process regulates the encoder to disregard user-identifiable patterns and tunes the decoder to shape the output independently of users in the training set. 

Another body of work~\cite{shredder, osia1, osia2, cloak} propose such schemes for computer vision tasks and preserving privacy of images.
Osia et al. propose Deep Private Feature Extraction (DPFE)~\cite{osia2} which aims at obfuscating input images to hinder the classification of given sensitive (private) labels, by modifying the network topology and re-training all the model parameters. 
DPFE partitions the network in two partitions, first partition to be deployed on the edge and the second on the cloud. It also modifies the network architecture by adding an auto-encoder in the middle and then re-training the entire network with its loss function.  The encoder part of the auto-encoder is deployed on the edge device, and the decoder is deployed on the server.
The auto-encoder aims to reduce the dimensions of the sent data which decreases the communication cost, alongside decreasing the amount of information that is sent, which helps the privacy. 

DPFE's loss function can be seen in Equation~\ref{eq:dpfe}. It is composed of three terms, first, the cross-entropy loss for a classification problem consisting of M classes ($y_{o,c}$ indicates whether the observation $o$ belongs to class $c$ and $p_{o,c}$ is the probability given by the network for the observation to belong to class $c$). This term aims at maintaining accuracy. Second, a term that tries to decrease the distance between intermediate activations of inputs with different private labels, and a final term which tries to increase the distance between intermediate activations of inputs with the same private label. $\gamma$ is a constant which depends on the number of dimensions and the training data, it is used as a normalization factor. $k$ is also a constant which depends on the training data.  $i$ and $j$ are iterators over the main batch and a random batch, respectively and $Y$ is the private label for that batch member. 
\begin{equation}
\begin{split}
    - \sum_{c=1}^{M} y_{o,c}log(p_{o,c}) 
    +\gamma( \sum_{(i,j):Y_i \neq Y_j}^{}||a'_i - a'_j||_2 \\
    +\sum_{(i,j):Y_i = Y_j}^{}(k-||a'_i - a'_j||_2))
\end{split}
\label{eq:dpfe}
\end{equation}

DPFE retrains the given neural network and the auto-encoder with this loss function. The training can be seen as an attempt to create clustered representations of data, where the inputs with the same private labels go in different clusters, and inputs with different labels are pushed to the same cluster, to mislead any adversary who tries to infer the private labels. 
Given its loss function, DPFE cannot operate without the private labels. Therefore, if there is a setting in which no sensitive labels are provided, DPFE cannot be used.
After training, for each inference request, a randomly generated noise is added to the intermediate results on the fly. This noise is not there to achieve differential privacy.

More recently, Mireshghallah et al. suggested Shredder~\cite{shredder}, a framework that without altering the topology or the weights of a pre-trained network, heuristically learns additive noise distributions that reduce the information content of communicated data while incurring minimal loss to the inference accuracy.
Shredder's approach also consists of cutting the neural network and executing a part of it on the edge device, similar to DPFE. This approach has been shown to decrease the overall execution time in some cases~\cite{shredder}, compared to running the entire neural network on the cloud, since the communication takes the bulk of time and sensing intermediate representations can sometimes save on the communication since there are fewer dimensions. 

Shredder initializes a noise tensor, with the same dimension as the intermediate activation, by sampling from a Laplace distribution with location of $0$, and scale of $b$, which is a hyperparameter. Then, using the loss function shown in  Equation~\ref{eq:shredder}, it tries to maintain the accuracy of the model (first term), while increasing the amount of additive noise (second term). $\lambda$ is a knob that provides an accuracy-privacy trade-off.
\begin{equation} \label{eq:shredder}
    - \sum_{c=1}^{M} y_{o,c}log(p_{o,c}) - \lambda\sum_{i=1}^{N}{|n_i|}
\end{equation}
Once the training is terminated, a Laplace distribution is fit to the trained tensor, and the parameters of that distribution, alongside the order of the elements, are saved. A collection of these distributions are gathered. During inference, noise is sampled from one of the saved distributions and re-ordered to match the saved order. This noise tensor is then added to the intermediate representation, before being sent to the cloud. 
Both DPFE and Shredder empirically demonstrate a reduction in the number of mutual information bits between the original data and the sent intermediate representation.  
DPFE can only be effective if the user knows what s/he wants to protect against, whereas Shredder offers a more general approach that tries to obliterate any information that is irrelevant to the primary task. Empirical evaluations showed that Shredder can in average loose more mutual information, compared to DPFE. However, in the task of inferring private labels, DPFE performs slightly better by causing a higher misclassification rate for the adversary since it has access to the private labels during training time.

\begin{figure}
    \vspace{-2ex}
    \centering
    \includegraphics[width=0.75\linewidth]{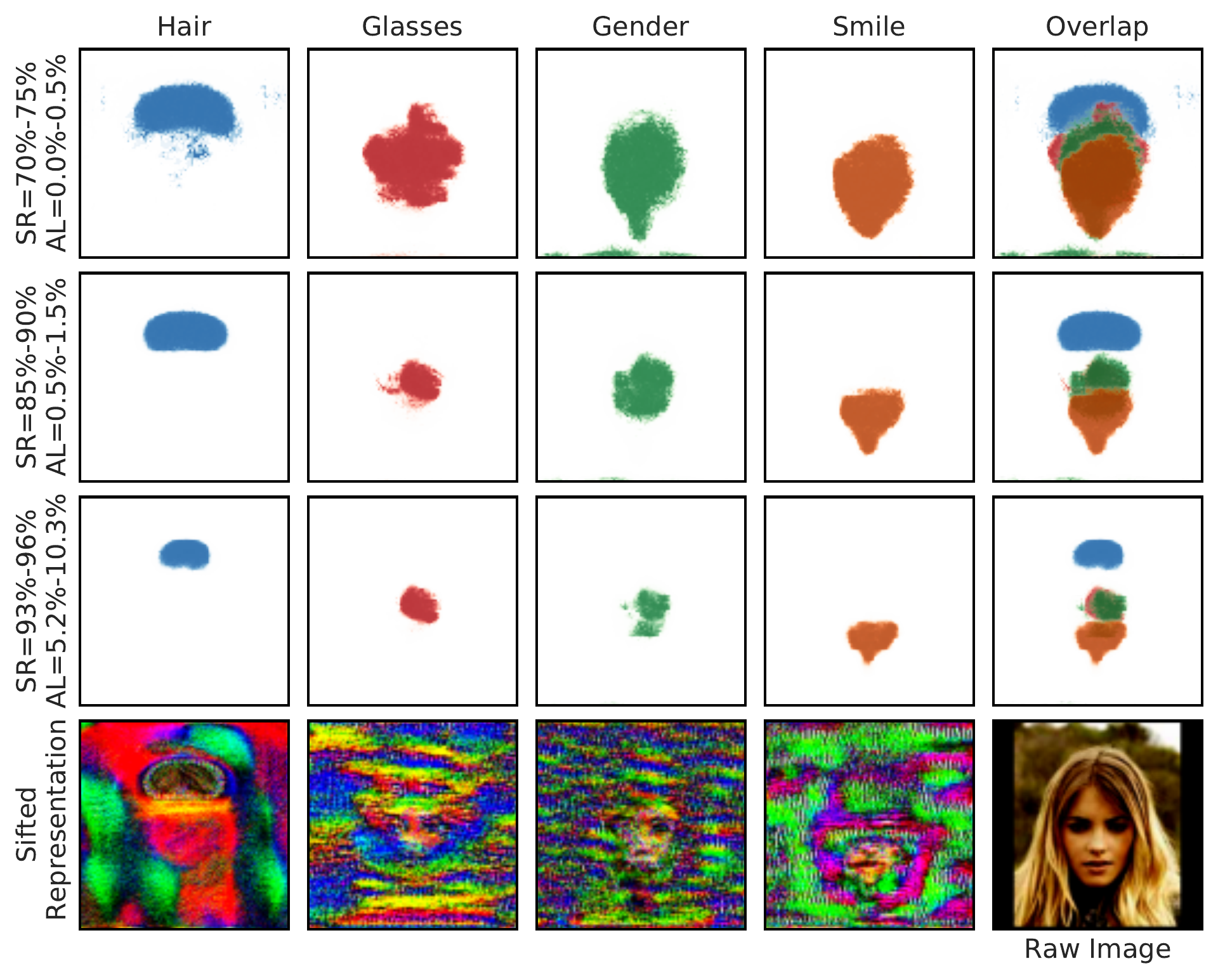}
    \caption{Cloak's discovered features for target DNN classifiers (VGG-16) for black-hair color, eyeglasses, gender, and smile detection. The colored features are conducive to the task. The 3 sets of features depicted for each task correspond to different suppression ratios (SR). AL denotes the range of accuracy loss imposed by the suppression.  
    }
    \label{fig:screen}
    \vspace{-2ex}
\end{figure}

More recently, Mireshghallah et al. propose a non-intrusive interpretable approach dubbed Cloak, in which there is no need to change/retrain the network parameters, nor partition it. 
This work attempts to explain the decision making process of deep neural networks (DNNs) by separating the subset of input features that are essential and unessential for the decisions made by the DNN during test-time inference. This separation is made possible through the adoption of information-theoretic bounds for  different set of features. 
%
%
After identifying the essential subset, Cloak suppresses the rest of the features using learned values, and only sends the essential ones. In this respect, Cloak offers an interpretable privacy-preserving mechanism.
An example of representations produced by Cloak and the conducive/non-conducive feature separation can be seen in Figure~\ref{fig:screen}.

\section{Privacy-Enhancing Execution Models and Environments}  \label{sec:models}

Apart from privacy-preserving schemes which are methods that directly optimize for a given definition of privacy, there are given execution models and environments that help enhance privacy and are not by themselves privacy-preserving.
In this section, we will briefly discuss federated learning, split learning and trusted execution environments, which have been used to enhance privacy. These methods are usually accompanied by privacy-preserving schemes from the previous section.

\subsection{Federated Learning}
\begin{figure}
    \centering
    \includegraphics[width=0.7\linewidth]{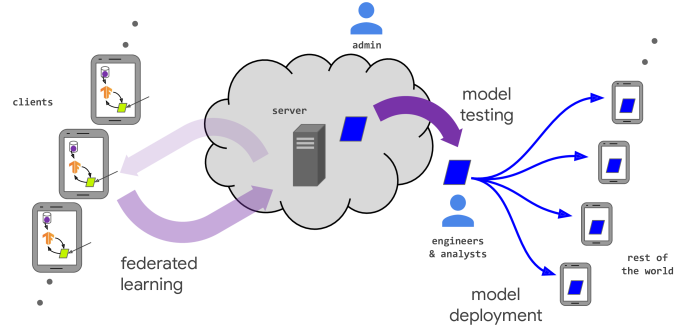}
    \caption{The workflow of federated learning model~\cite{fdsurvey}.}
    \label{fig:federated}
\end{figure}
Federated learning (FL) is a machine learning setting where many clients collaboratively train a model under the administration of a central server while keeping the training data local. 
Federated learning is built on the principles of focused collection and data minimization which can alleviate the privacy risks of centralized machine learning~\cite{fdsurvey}. 
The workflow of federated learning can be seen in Figure~\ref{fig:federated}. This workflow is broken into six stages~\cite{fdsurvey}:

\begin{enumerate}
    \item Problem identification: The problem that is to be solved using federated learning should first be defined.
    \item Client instrumentation: The clients can be instructed to save the data needed for training. For example, the applications running on the edge devices might need to locally save some metadata (e.g. user interaction data) alongside the main data (for instance text messages). 
    \item Simulation prototyping (optional): The engineer who is deploying the system might need to prototype different architectures and try different hyperparameters in a federated learning simulation.
    \item Federated model training: Multiple federated training tasks are initiated which train different variations of the model or use different optimization hyperparameters.
    \item Model evaluation: When the tasks are done with the training phase (usually after a few days), the models are analyzed and evaluated, either on standard centralized datasets or on local client data. 
    \item Deployment: When the analysis is finished and a model is selected, the launch process is initiated. This process consists of live A/B testing, manual quality assurance, and a staged roll-out. 

\end{enumerate}{}

Federated learning is being widely used with SMC and differential Privacy~\cite{fdsurvey, smc-fed,GANFL}.Bonawitz et al. apply \textbf{Secure aggregation}
to privately combine the outputs of local machine learning on user devices in the federated learning setup, to update a global model.
Secure aggregation refers to the computation of a sum in a multiparty setting, where no party reveals its update in the clear, even to the aggregator.
When Secure Aggregation is added to Federated Learning, the aggregation of
model updates is performed by a virtual incorruptible third party induced by secure multiparty communication. With this setup, the cloud provider learns only the aggregated model update.
There are also bodies of work that consider shuffling of user data, so as to hide the origin of each data item.The works of Cheu et al., and Balle et al. have proposed secure aggregation protocols that satisfy differential privacy guarantees in the shuffle model~\cite{shuffle1, shuffle2}. More recent work~\cite{shuffle3} mitigates the incurred error and communication overheads in shuffle model. 
More in-depth details of federated learning workflow and integration is out of the scope of this survey.

\subsection{Split Learning} 
\begin{figure}
\begin{subfigure}{.49\linewidth}
  \centering
  \includegraphics[width=.68\linewidth]{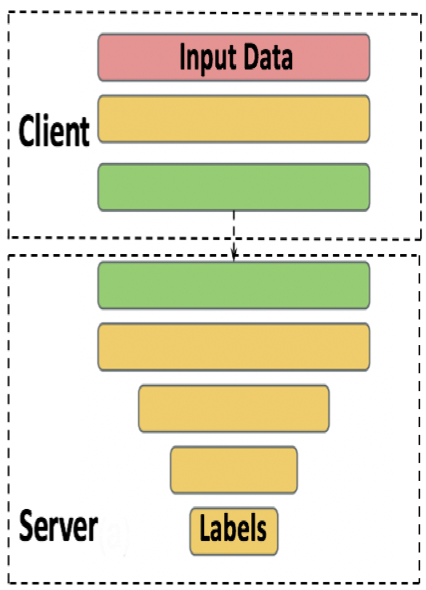}  
  \caption{Vanilla split learning}
  \label{man}
\end{subfigure}
\begin{subfigure}{.49\linewidth}
  \centering
  \includegraphics[width=.68\linewidth]{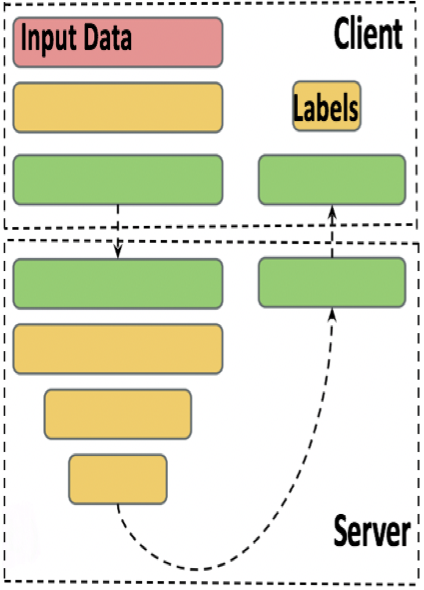}  
  \caption{Boomerang split learning}
  \label{woman}
\end{subfigure}
\caption{The vanilla configuration of split learning where raw data is not shared between client and server, and boomerang (U-shaped) configuration where neither raw data nor  the labels are shared between client and server~\cite{split-image}.}
 \label{fig:sl}
\end{figure}
Split-learning is an execution model where the neural network is split, between the client and the server~\cite{main-split}. This is very similar to the neural network partitioning described in Shredder~\cite{shredder} and DPFE~\cite{osia2}. 
Vanilla split learning is formed by each client computing the forward pass through a deep neural network up to a specific layer, called the cut layer.
The outputs of the cut layer, referred to as smashed data, are sent from the edge device to another entity (either the server or another client), which completes the rest of the computation. With this execution scheme, a round of forward pass is computed without sharing raw data. The gradients can then be backpropagated from the server to the cut layer in a similar fashion. The gradients at the cut layer are transferred back to the clients, where the rest of the backpropagation is completed. In this fashion, the training or inference is done without having clients directly access each other's raw data.  An instantiation of this setup where labels are also not shared along with raw data is shown in Figure~\ref{fig:sl}.


\subsection{Trusted Execution Environments (TEEs)}

Trusted execution environments, also referred to as secure enclaves, provide opportunities to move parts of decentralized learning or inference processes into a trusted environment in the cloud, whose code can be attested and verified. Recently, Mo et al. have suggested a framework that uses an edge device’s Trusted Execution Environment (TEE) in conjunction with model partitioning to limit the attack surface against DNNs~\cite{Mo2020DarkneTZTM}.
TEEs can provide integrity and confidentiality during execution. TEEs have been deployed in many forms, including Intel’s SGX-enabled CPUs~\cite{intelsgx, murali, Hashemi2020DarKnightAD, tramer2018slalom}, Arm’s TrustZone~\cite{armtrust}.
This execution model, however, requires the users to send their data to an enclave running on remote servers which allows the remote server to have access to the raw data and as the new breaches in hardware~\cite{spectre,meltdown,l1tf,mds,packetchasing,csf} show, the access can lead to comprised privacy.
\vspace{-2ex}
\section{Conclusion}

The surge in the use of machine learning is due to the growth in data and compute. 
The data mostly comes from people~\cite{nytimes2} and includes an abundance of sensitive information. 
This work tries to provide a comprehensive and systematic summary of the efforts made to protect privacy of users in deep learning settings.
We find an apparent disparity in the number of efforts between data aggregation, training, and inference phases. 
In particular, little attention has been made to privacy of the users during inference phase. 


\bibliography{paper}
\bibliographystyle{ieeetr}

\end{document}